\documentclass[9.5pt,journal,compsoc, cspaper]{IEEEtran}

\usepackage{graphicx}
\usepackage[nocompress]{cite}
\usepackage{algorithm}
\usepackage[noend]{algorithmic}
\usepackage{fixltx2e}
\usepackage{multirow}

\newcommand\Tstrut{\rule{0pt}{2.1ex}}         
\newcommand\Bstrut{\rule[-0.8ex]{0pt}{0pt}}   

\usepackage{flushend}
\newcommand{\subparagraph}{}

\usepackage[noindentafter]{titlesec}
\titlespacing\section{0pt}{12pt}{3pt}
\titlespacing\subsection{0pt}{10pt}{3pt}

\begin{document}

\title{EDCompress: Energy-Aware Model Compression for Dataflows}

\author{Zhehui~Wang,
        Tao~Luo,
        ~Joey~Tianyi~Zhou,
        and ~Rick~Siow~Mong~Goh
\IEEEcompsocitemizethanks{\IEEEcompsocthanksitem Z. Wang, T. Luo, J. Zhou and R. Goh are with the Institute of High Performance Computing, Agency for Science, Technology and Research (A*STAR), Singapore.\protect\\
E-mail:\{wang\_zhehui, luo\_tao, joey\_zhou, gohsm\}@ihpc.a-star.edu.sg
}}

\IEEEtitleabstractindextext{%
\begin{abstract}
Edge devices demand low energy consumption, cost and small form factor. To efficiently deploy convolutional neural network (CNN) models on edge device, energy-aware model compression becomes extremely important. However, existing work did not study this problem well because the lack of considering the diversity of dataflow types in hardware architectures. In this paper, we propose EDCompress, an Energy-aware model compression method for various Dataflows. It can effectively reduce the energy consumption of various edge devices, with different dataflow types. Considering the very nature of model compression procedures, we recast the optimization process to a multi-step problem, and solve it by reinforcement learning algorithms. Experiments show that EDCompress could improve 20X, 17X, 37X energy efficiency in VGG-16, MobileNet, LeNet-5 networks, respectively, with negligible loss of accuracy. EDCompress could also find the optimal dataflow type for specific neural networks in terms of energy consumption, which can guide the deployment of CNN models on hardware systems.
\end{abstract}

\begin{IEEEkeywords}
Efficient AI, Edge Device, Energy-Aware, Model Compression, Dataflow
\end{IEEEkeywords}

}

%
\maketitle

\IEEEdisplaynontitleabstractindextext

%
\IEEEpeerreviewmaketitle

\IEEEraisesectionheading{\section{Introduction}\label{sec:introduction}}


\IEEEPARstart{C}{onvolutional} neural network (CNN) shows good performance in various applications such as image classification and object detection. However, traditional CNN is in large scale, which makes it challenging to implement on edge devices. For example, the VGG-16 network contains 528 MB weights~\cite{simonyan2014very}. To classify one image, we need to perform $1.5\times 10^{10}$ multiply–accumulate (MAC) operations. There are two consequences. First, the limited memory space of edge devices cannot store the parameters. Second, the edge device becomes power hungry because the calculation and data movement operations consume a large amount of energy. 

\begin{figure}[!t]
  \centering
  \includegraphics[width=3.5 in]{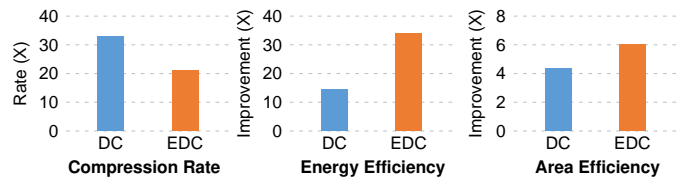}\\
  \caption{Comparison between our EDCompress (EDC) and Deep Compression (DC) in terms of the improvements of energy efficiency and area efficiency, the larger the better}
  \label{f:compare1}
\end{figure}

Model compression method such as quantization and pruning, is an emerging technique developed in recent years to alleviate this problem. Previous model compression methods target on the reduction of model size. For example, Han~\emph{et al.} proposed Deep Compressing method~\cite{han2015deep}, which helps to fit the neural networks into the on-chip memory of edge devices. However, a neural network with reduced model size does not guarantee that it is also energy efficient. To prove this, we compare our work EDCompress (EDC) with Deep Compression (DC) in Figure~\ref{f:compare1}. We can see that although EDCompress shows lower compression rate, it has higher energy and area efficiency than DC.

The energy consumption of edge devices is highly related to its dataflow design~\cite{yang2018dnn}. We experimentally observe that a large portion of the energy is consumed on the data movement (e.g., around 72\% in VGG-16) in computing the convolution layers. 
In the past few years, many dataflow types were developed to reuse the data and thus substantially improve the energy efficiency of the accelerators (e.g., ~\cite{du2015shidiannao},~\cite{qiu2016going},~\cite{chen2016eyeriss},~\cite{chen2014diannao}, etc.). These dataflow types use different policies to reuse the data, and therefore have different impacts on energy consumption.
Unfortunately, existing model compression work did not consider the diversity of dataflow types in hardware architectures, leading to sub-optimal results.

In this paper, we propose \textbf{EDCompress}, an \textbf{E}nergy-aware model \textbf{compress}ion method for various \textbf{D}ataflows. Compared with previous work, it has two different features:
\begin{itemize}
  \item \textbf{Hardware awareness}: To our best knowledge, this is the first paper to study the impact of the dataflow types on energy consumption in model compression. We propose the model compression method which can be adapted to different hardware systems with different dataflow types.
  \item \textbf{Automated Approach}: We first formulate the energy-aware model compression as a multi-step optimization problem. At each step, we partially quantize or prune the model. We further recast it into a reinforcement learning problem, and enable automatic search of the best model compression strategy.
\end{itemize}
 
\section{Related Work}

Traditional model compressing target on the reduction of model size. There are three major methods: pruning, quantization, and weight sharing. Most of the model compression work forced on pruning, where we replace those weights with small absolute values by zeros~\cite{he2018amc}\cite{cai2018proxylessnas}\cite{yang2018netadapt}\cite{guo2016dynamic}\cite{xiao2017building}\cite{liu2018frequency}\cite{chang2018prune}\cite{manessi2018automated}\cite{li2016pruning}\cite{singh2019play}\cite{kang2019accelerator}\cite{lemaire2019structured}\cite{8781626}\cite{fang2018nestdnn}\cite{yang2018energy}\cite{hacene2018quantized}.
Some work focused on quantization, where we decrease the precision of the weights and the activations~\cite{ding2019req}\cite{geng2019dataflow}\cite{8578919}. Other work focused on weight sharing, where we cluster all the weights in the filter, and only store the index and the centroids in the memory~\cite{xiao2017building}\cite{han2016eie}\cite{ullrich2017soft}.

Recently, edge devices are becoming increasing popular for AI applications. However, considering the large amount of energy consumed in inference, the deployment of traditional compression methods on edge devices becomes  infeasible. To solve this problem, there are only a few research works which start to study energy-aware model compression recently.
Wang~\emph{et al.}~\cite{wang2019haq}  first manually set a constraint of energy consumption and then explore the quantization policy under this predefined constraint. Yang~\emph{et al.}~\cite{yang2017designing} developed a heuristic algorithm to prune the model. They search layers which consume most of the energy consumption, and pruning those layers in priority. In their optimization process, the energy consumption is a indirect target. In contrast with these two methods, we directly take energy consumption as one of targets to optimize the model and seek a better trade-off between the accuracy, energy and model size. In addition, to our best knowledge,  it is the first paper to consider the impact of dataflow types on energy consumption in compressing models.

\begin{algorithm}[!t]
\begin{algorithmic}
\FOR{$c_o$ in range ($C_O$)}
    \FOR{$c_i$ in range ($C_I$)}
        \FOR{$x$ in range ($X$)}
            \FOR{$y$ in range ($Y$)}
                \FOR{$f_x$ from -($F_X$-1)/2 to ($F_X$-1)/2}
                    \FOR{$f_y$ from -($F_Y$-1)/2 to ($F_Y$-1)/2}
                        \STATE $O$[$c_o$][x][y]+= \\                       $I$[$c_i$][x+$f_x$][y+$f_y$]$\times$$W$[$c_o$][$c_i$][$f_x$][$f_y$]
                    \ENDFOR
                \ENDFOR
            \ENDFOR
        \ENDFOR
    \ENDFOR
\ENDFOR
\end{algorithmic}
\caption{Computation of a typical convolutional layer}
\label{conv}
\end{algorithm}

\renewcommand{\arraystretch}{0.8}
\begin{table}[!t]
\caption{Popular dataflow types}
\vspace{-5pt}
 \resizebox{1.00\linewidth}{!}{ \begin{tabular}{c|c|c|c}
\hline
 Dataflow    &   Applied by     &  Dataflow    &    Applied by   \Tstrut \Bstrut\\
\hline
$X:Y$   &  ~\cite{du2015shidiannao}~\cite{song2018towards}           & $F_X:F_Y$    & ~\cite{qiu2016going}      \Tstrut\\
$X:F_X$  & ~\cite{chen2016eyeriss}~\cite{gao2017tetris}~\cite{li2016high}            & $C_I:C_O$    &~\cite{chen2014diannao}~\cite{jouppi2017datacenter}~\cite{zhang2015optimizing}~\cite{alwani2016fused}~\cite{shen2016overcoming}~\cite{suda2016throughput}        \Bstrut\\
\hline
\end{tabular}}
\label{datatype}
\end{table}

\section{Quantization/Pruning for Dataflows}

 

Dataflow can be considered as the mapping strategy between the mathematical operations and the processing elements~\cite{yang2018dnn}. Algorithm~\ref{conv} shows the computation of a typical convolutional layer. The algorithm contains six loops. One loop corresponds to one dimension in either the filter or the feature map. Here, $C_O$ and $C_I$ denote the number of output and input channels. $X$ and $Y$ denote the width and height of the feature map. $F_X$ and $F_y$ denote the width and height of the filter.
In each iteration of the innermost loop, we perform a basic operation called multiply–accumulate (MAC). Before the MAC operation, we read three elements from the memory, one from the input feature map, one from the weight, and one from the output feature map. After the MAC operation, we write the result into the memory. Most of the energy is spent on the MAC calculation and data movement. To compute one conventional layer, we need to execute $C_O\cdot C_I\cdot X\cdot Y\cdot F_X\cdot F_Y$ MAC operations in total.

\begin{figure}[!t]
  \centering
  \includegraphics[width=3.5in ]{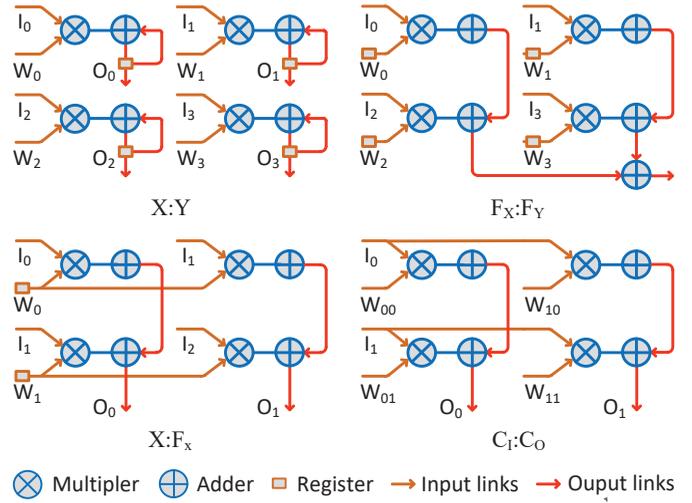}\\
  \caption{The hardware accelerators with four popular dataflow types. $W_k/W_{kk}$, $I_k$, and $O_k$ are each element in the weight, the input feature map, and the output feature map}
  \label{f:optimizestep}
\end{figure}



In edge devices, there is an array or a matrix of processing elements, each one can execute the MAC operation independently. The strategy to map the operation into those elements becomes a key consideration in the hardware. This is a large design space to explore. For example, given an array of processing elements, we can unroll any one of the loops in the algorithm, and map each iteration in the loop into each processing element in the array. By similar rules, we can further unroll two loops in the algorithm and map the MAC operations into a matrix of processing elements. With six loops in total, there are $C_6^2$=15 possibilities, each one corresponds to one dataflow. Here, we introduce four popular dataflow types in Table~\ref{datatype}. They are denoted as A:B, where A and B stand for the name of each loop.

Different dataflow designs employ different data movement policies, resuslt in different energy consumption. In Figure~\ref{f:optimizestep}, we show four popular dataflow types. To simplify the figure, we only put four processing elements in each example. In real implementations, the $A$:$B$ dataflow design requires $A\cdot B$ processing elements. In $X$:$Y$, we store MAC operation results in registers at output ports of processing elements. At each iteration, we read the last MAC operation result from registers. In $F_X$:$F_Y$, we store $F_X\cdot F_Y$ weights in registers at input ports of processing elements. At each iteration, we sum up $F_X\cdot F_Y$ MAC operation results. In $X$:$F_x$, we store $F_X$ weights in registers at input ports of processing elements. At each iteration, we reuse the weights by $X$ times, and sum up $F_X$ MAC operation results. In $C_I$:$C_O$, at each iteration, we reuse the input feature map by $C_O$ times, and sum up $C_I$ MAC operation results.


\begin{figure}[!t]
  \centering
  \includegraphics[width=3.5in]{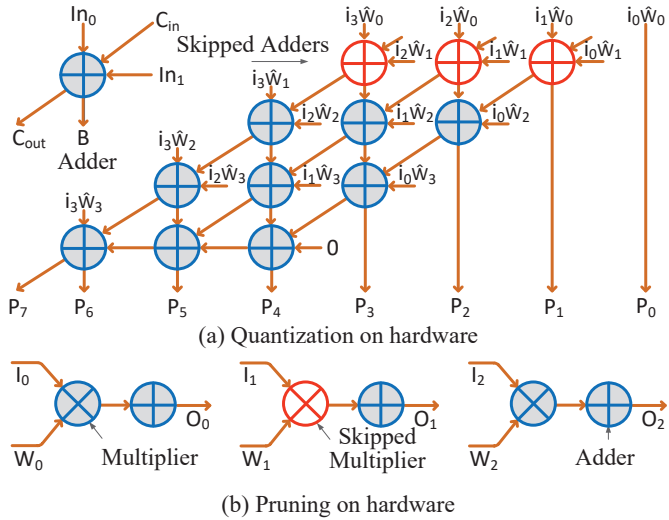}\\
  \caption{(a) If the weights are quantized from 4 bits to 3 bits, we can skip the first row of adders ; (b) If the weights are pruned, we can skip those multipliers whose weights are zeros; $I_k$ and $W_k$ are each element in the input feature map and the weight, $i_k$ and $\hat{w}_k$ are each bit in $I_k$ and $W_k$ }
  \label{f:mapchoice}
\end{figure}

\subsection{Improvement on Energy Efficiency}

We use quantization and pruning to compress the model because they show better performance over weight sharing~\cite{han2015deep}.
To quantize a model, we lower the precision of parameters based on the quantization depth (the number of digits presenting a parameter). After quantization, the low-precision parameters are still expected to store sufficient information for  inference. To prune a model, we replace some of the parameters in the model by zeros. A well-trained model usually contains many weights with negligible values. We sort all the weights in the filter, and replace those weights with the small absolute values by zeros.

We can save energy of the logic circuits using quantization and pruning. Figure~\ref{f:mapchoice} (a) shows the inner structure of a 4 bits$\times$4 bits multiplier, which contains 12 adders. If the weights are quantized from 4 bits to 3 bits, we can skip the last row of adders, and thus save the energy consumption. In real applications, a high-precision model with 32FP data type (32 bit float point) requires 23 bit$\times$23 bit multipliers, with 506 adders in total. If both the activations and weights can be quantized, we can save a plenty of energy. For example, if the activations are quantized from 32FP to 16FP, and the weights are quantized from 32FP to 8INT (8 bit integer), only 10 bit$\times$8 bit multipliers are required, with 72 adders in total, which is 86\% less than the original amount. Figure~\ref{f:mapchoice} (b) shows an array of three processing elements, each containing a multiplier and an adder. If the weights are pruned, some processing elements would have inputs equaling zero. In this case, we can skip the corresponding multiplier, and save the energy consumption.

We can also save energy of the memory modules using quantization and pruning. To classify an image, we need to store all the weights, and put the intermediate feature map of each layer into the memory. The memory can be either the on-chip memory or the off-chip memory. No matter which type of memory we use, the data movement energy consumption of memory modules are proportional to the total amount of data transmitted, counted in bits. To decrease this value, we can either reduce the size of parameters by quantization, or reduce the number of parameters by pruning. For example, if we quantize the parameters from 32FP to 16FP, and prune half of the parameters, then roughly 75\% of the energy of memory modules can be reduced.

\begin{figure}[!t]
  \centering
  \includegraphics[width=3.5in ]{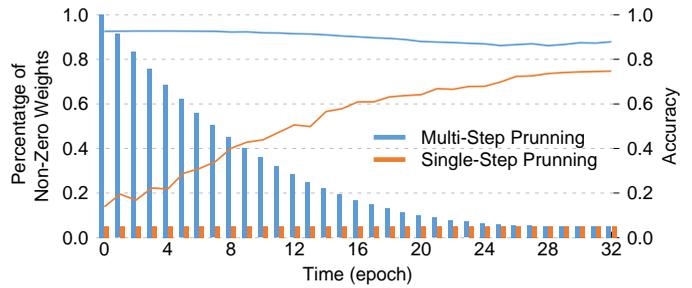}\\
  \caption{Comparison between multi-step pruning and one-step pruning, tested on CIFAR-10 dataset using VGG-16. The boxes denote the percentage of non-zero weights, and the curves denote the accuracy}
  \label{f:multistep}
\end{figure}

\subsection{Recasting to the Multi-Step Problem}

We recast the model compression process to a multi-step problem. Our goal is to lower the energy consumption of edge devices while keeping the accuracy of the model. Instead of quantizing/pruning the model directly in one step, our final target is approached through a sequence of quantization/pruning steps. This is because we cannot alter the parameters too much at one time. Otherwise, the performance of the model will be reduced obviously, and it will be too difficult to restore the model~\cite{zhu2017prune}. Figure~\ref{f:multistep} shows the comparison between the multi-step pruning and the single-step pruning. We test CIFAR-10 dataset from a well-trained VGG-16 model. For the multi-step pruning, we gradually decrease the percentage of non-zero weights from 100\% to 5\% in 32 steps. After each step, we re-train the model by one epoch. For the single-step pruning, the percentage of non-zero weights is kept on 5\%, and we re-train the model by 32 epochs. From the figure we can see that the multi-step pruning can achieve better accuracy than the single-step pruning.

\begin{figure}[!t]
  \centering
  \includegraphics[width=3.5in]{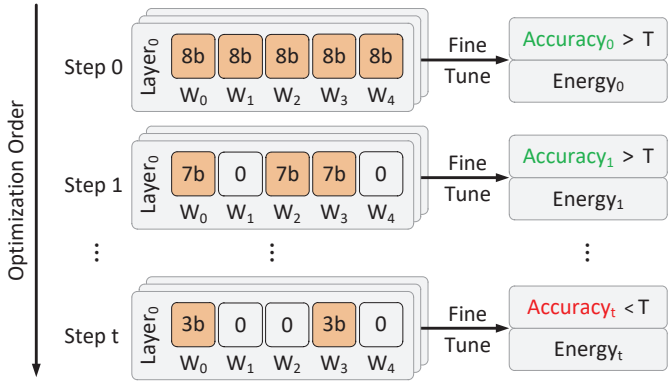}\\
  \caption{The multi-step optimization process, where the final quantization depth and pruning amount are approached gradually}
  \label{f:processorder}
\end{figure}

We show an example of the multi-step optimization process in Figure~\ref{f:processorder}. In each step, we increase or decrease the quantization depth (the precision of the parameters) or the prune amount in each layer. For example, in step 1, we prune 40\% weights, and the left weights are quantized by 7 bits. We then fine tune the model, train a few more epochs, and check the accuracy and energy of the model. If the accuracy is greater than threshold, we change the quantization depth and the pruning amount, and repeat the optimization process. In step $t$, we prune 60\% weights, and quantize the remaining weights by 3 bit. Since the model accuracy drops a lot at this step, we stop the optimization process.

The searching space of optimal solutions in this problem is very huge for humans. Manually optimizing the hardware accelerators would become a tough task considering the diversity of dataflow designs and quantization/pruning choices. In total, there are 15 different dataflow types. The parameters in each layer can be quantized from 23 bit to 1 bit, and the pruning amount in each layer can range from 0\% to 100\%. In general, an $L$-layer model has $15\times100^L\times23^L$ possible choices, assuming 1\% pruning amount granularity. Given such large design space, engineers would face many choices in compressing the model. Hence, developing automated optimizing approach considering the characteristics of dataflow becomes critical for hardware accelerators.

\subsection{Optimization through Reinforcement Learning}
\begin{figure}[!t]
  \centering
  \includegraphics[width=3.5in]{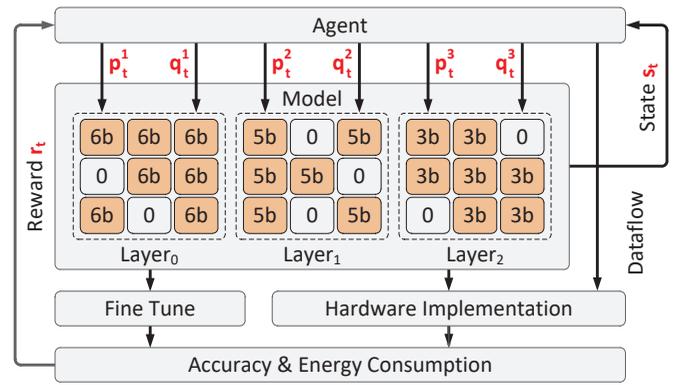}\\
  \caption{The reinforcement learning based optimization model. The agent increases or decreases the quantization depth/pruning remaining amount at each step}
  \label{f:overview}
\end{figure}

Considering the nature of multi-step problem, we propose to use reinforcement learning  to  search the best model compression strategy with high energy efficiency.   The reinforcement learning allows us to  automatically explore the design space, and find the optimal quantization/pruning policies for each dataflow on the hardware. We show the overview of our reinforcement learning model in Figure~\ref{f:overview}. Specifically, in each episode, an agent interacts with the environment (the CNN model) via a sequence of steps. In each step $t$, the agent generates an action vector $a_t$ based on the state vector of the environment $S_t$. The environment responds to action $a_t$, quantize/prune the parameters in the model, and change its state to $S_{t+1}$. The model is then fine-tuned by one or few epochs, and a reward $r_t$ considering both accuracy and energy consumption is returned. For large dataset such as ImageNet, the model is not fine-tuned in the first few steps. The agent then updates its own parameters for achieving higher rewards in later actions. In each episode, we start from 100\% pruning remaining amount and 8-bit quantization depth. An episode ends if the number of steps exceeds the limit, or the accuracy of the model drops below the predefined threshold. Equation~\ref{e:qp} models the quantization depth and the pruning remaining amount.
\begin{equation}
Q_t^l=Q_0^l+\sum_{i=0}^{t-1}q_i^l\gamma^i \ \ \ \ \ \ \ \ \ \ \ \ P_t^l=P_0^l+\sum_{i=0}^{t-1}p_i^l\gamma^i
\label{e:qp}
\end{equation}
 Here, $Q_0^l$ and $P_0^l$ denote the original quantization depth and pruning remaining amount of $l$-th layer in the CNN model before the optimization.
$Q_t^l$ and $P_t^l$ denote the quantization depth and the pruning remaining amount after optimization step $t-1$ ($t\geq1$). To obtain $Q_t^l$ and $P_t^l$, we need $t$ steps of optimization. In step $i$, the agent changes the values of $Q^l$ and $P^l$ by $q_i^l$ and $p_i^l$ respectively. To get a better optimization result, we take smaller steps when $Q_t^l$ and $P_t^l$ are close to the optimal point. The discount factor $\gamma$ is used to regulate the variance of $q_i^l$ and $p_i^l$. We test different values of $\gamma$ in experiments, and find that $\gamma=0.9$ is an optimal value.

\begin{equation}
a_t=(\bigcup_{l=0}^{L-1}\{q_t^l\})\cup(\bigcup_{l=0}^{L-1}\{p_t^l\})
\label{e:at}
\end{equation}

The action $a_t$ can be expressed by Equation~\ref{e:at}. Here $a_t$ is the set containing changes of $Q$ and $P$ in all layers. Although the quantization depth is a discrete variable, we use the continuous action space. This is because we don't want to lose the small changes of the quantization depth accumulated in each optimization step. When we fine-tune the network, we round the quantization depth to its nearest integer value.
\begin{equation}
s_t=(\bigcup_{t-\tau}^{m=t}\bigcup_{l=0}^{L-1}\{Q_m^l\})\cup(\bigcup_{t-\tau}^{m=t}\bigcup_{l=0}^{L-1}\{P_m^l\})\cup(\bigcup_{t-\tau}^{m=t}\{r_m\})\cup\{t\}
\label{e:st}
\end{equation}

\renewcommand{\arraystretch}{0.60}

\begin{table*}[!t]
\begin{minipage}{.45\linewidth}
\vspace{0pt}
\caption{EDCompress and HAQ~\cite{wang2019haq} on ImageNet using MobileNet}
\vspace{-5pt}
  \centering
 \resizebox{1.00\linewidth}{!}{ \begin{tabular}{c  c  c  c  c}
     \hline
       \multirow{2}{*}{Dataflow}       &  \multicolumn{2}{c}{Norm. Energy}  &\multicolumn{2}{c}{Norm. Area}\Tstrut\Bstrut\\
             & ~\cite{wang2019haq}  &Ours                      &~\cite{wang2019haq}  &Ours\Bstrut\\
        \hline
        $X:Y$    &5.44 &\textbf{1.41} &26.1    &\textbf{5.27}\Tstrut\\
        $F_X:F_Y$ &6.31 &\textbf{1.81} &2.53  &\textbf{1.00}  \\
        $X:F_X$ &6.32  &\textbf{1.81} &2.53   &\textbf{1.00}  \\
        $C_I:C_O$ &4.48  &\textbf{1.00} &505  &\textbf{92.0}\Bstrut\\
        \hline
        \ \ \ \ \ \ Top-1 Accuracy\ \ \ \ \ \ &64.8 &68.3 &64.8 &68.3\Tstrut\\
        \ \ \ \ \ \ Top-5 Accuracy\ \ \ \ \ \ &85.9 &88.3 &85.9 &88.3\Bstrut\\
         \hline
 \label{t:Compare21}
    \end{tabular}}
\end{minipage}
\hspace{1.8em} 
\begin{minipage}{.52\linewidth}
\vspace{0pt}
\caption{EDCompress and the previous work~\cite{li2016pruning}~\cite{singh2019play} on CIFAR-10 using VGG-16 \newline }
\vspace{-5pt}
  \centering
 \resizebox{1.00\linewidth}{!}{ \begin{tabular}{c  c c c  c c c}
     \hline
       \multirow{2}{*}{Dataflow}       &  \multicolumn{3}{c}{Norm. Energy}  &\multicolumn{3}{c}{Norm. Area}\Tstrut\Bstrut\\
             &~\cite{li2016pruning} &~\cite{singh2019play} &Ours                      &~\cite{li2016pruning} &~\cite{singh2019play}  &Ours\Bstrut\\
\hline
$X:Y$    &24.41 &15.10 &\textbf{1.69} &7.78 &5.56   &\textbf{1.00}\Tstrut\\
$F_X:F_Y$ &22.61 &14.42 &\textbf{2.31} &6.42 &4.20 &\textbf{1.27}  \\
$X:F_X$ &22.17 &15.10 &\textbf{2.73} &6.42 &4.20 &\textbf{1.42}  \\
$C_I:C_O$ &19.68  &12.21 &\textbf{1.00} &434 &431 &\textbf{47.58}\Bstrut\\
\hline
\ \ \ Accuracy \ \ \  &93.1  &93.4  &91.3  &93.1  &93.4  &91.3\Tstrut\Bstrut\\
 \hline
 \label{t:Compare22}
    \end{tabular} }
 \end{minipage}
\end{table*}

\renewcommand{\arraystretch}{0.80}
\begin{table*}[!t]
\caption{EDCompress and the previous work on MNIST using LeNet-5, due to memory module reuse, the total area is not the summation of sub-areas}
  \vspace{-5pt}
  \centering
   \resizebox{1.00\linewidth}{!}{ \begin{tabular}{ c c c c c c c c  c  c c c c c c  c }
     \hline
            \multicolumn{2}{c}{ } & \multicolumn{7}{c}{Energy ($\mu$J)}&\multicolumn{7}{c}{Area (mm$^2$)}\Tstrut\Bstrut\Bstrut\\
         \multicolumn{2}{c}{ } &~\cite{han2015deep} &~\cite{guo2016dynamic} &~\cite{xiao2017building} &~\cite{liu2018frequency}&~\cite{chang2018prune} &~\cite{manessi2018automated}& Ours &~\cite{han2015deep} &~\cite{guo2016dynamic} &~\cite{xiao2017building}&~\cite{liu2018frequency}&~\cite{chang2018prune} &~\cite{manessi2018automated} &Ours\Bstrut\\
 \hline
 \multirow{4}{*}{\rotatebox[origin=c]{90}{X:Y\ \ \  }}
&Conv1 &1.62 &3.34 &6.29 &2.93 &3.61 &15.76 &0.27 &0.95 &5.77 &5.77 &5.77 &5.77 &5.81 &0.53\Tstrut\\
&Conv2  &0.60 &1.47 &1.75 &1.20 &0.92 &8.29 &0.57 &0.15 &0.78 &0.78 &0.78 &0.78 &0.81 &0.09 \\
&FC1  &0.06 &0.07 &0.04 &0.06 &0.02 &0.32 &0.11 &0.02 &0.03 &0.03 &0.03 &0.03 &0.06 &0.02 \\
&FC2    &0.03 &0.09 &0.17 &0.07 &0.08 &1.14 &0.02 &0.08 &0.63 &0.63 &0.62 &0.62 &0.66 &0.07 \\
&Total		&2.31	&4.96	&8.25	&4.25	&4.62	&25.5	&\textbf{0.96}	&0.97	&5.81	&5.81	&5.80	&5.80	&5.83	&\textbf{0.55}\Bstrut\\
\hline
  \multirow{4}{*}{\rotatebox[origin=c]{90}{$F_X$:$F_Y$\ \ \  }}
&Conv1  &1.33 &3.09 &5.67 &2.73 &3.33 &13.91 &0.22 &0.05 &0.20 &0.20 &0.20 &0.20 &0.24 &0.03\Tstrut\\
&Conv2 &0.58 &1.58 &1.86 &1.29 &0.99 &7.78 &0.36 &0.06 &0.23 &0.23 &0.23 &0.23 &0.26 &0.04 \\
&FC1  &0.08 &0.08 &0.05 &0.07 &0.02 &0.38 &0.09 &0.04 &0.20 &0.21 &0.20 &0.20 &0.24 &0.03 \\
&FC2     &0.03 &0.09 &0.17 &0.07 &0.08 &1.14 &0.02 &0.08 &0.63 &0.63 &0.62 &0.62 &0.66 &0.06 \\
&Total		&2.03	&4.84	&7.75	&4.16	&4.42	&23.21	&\textbf{0.69}	&0.09	&0.66	&0.66	&0.66	&0.66	&0.7	&\textbf{0.08}\Bstrut\\
 \hline
\multirow{4}{*}{\rotatebox[origin=c]{90}{X:$F_X$\ \ \ }}
&Conv1  &1.17 &3.44 &6.05 &3.05 &3.70 &12.93 &0.39 &0.18 &1.04 &1.05 &1.04 &1.04 &1.08 &0.11\Tstrut\\
&Conv2  &0.71 &1.69 &2.00 &1.37 &1.04 &8.66 &0.53 &0.09 &0.41 &0.41 &0.41 &0.41 &0.45 &0.06 \\
&FC1  &0.10 &0.09 &0.05 &0.07 &0.02 &0.41 &0.20 &0.02 &0.06 &0.06 &0.06 &0.06 &0.09 &0.02 \\
&FC2     &0.03 &0.09 &0.17 &0.07 &0.08 &1.14 &0.02 &0.08 &0.63 &0.63 &0.62 &0.62 &0.66 &0.07 \\
&Total	&2.01	&5.31	&8.28	&4.56	&4.84	&23.13	&\textbf{1.14}	&0.2	&1.07	&1.07	&1.07	&1.07	&1.11	&\textbf{0.12}\Bstrut\\
\hline
\multirow{4}{*}{\rotatebox[origin=c]{90}{$C_I$:$C_O$\ \ \  }}
&Conv1  &2.08 &4.07 &7.58 &3.57 &4.40 &18.32 &0.36 &0.02 &0.06 &0.06 &0.06 &0.06 &0.10 &0.02\Tstrut\\
&Conv2  &0.73 &1.81 &2.14 &1.47 &1.13 &8.88 &0.63 &0.14 &0.75 &0.75 &0.75 &0.75 &0.78 &0.09 \\
&FC1  &0.06 &0.09 &0.06 &0.08 &0.03 &0.35 &0.08 &1.55 &14.11 &14.11 &14.11 &14.11 &14.15 &1.29 \\
&FC2     &0.03 &0.09 &0.17 &0.07 &0.08 &1.14 &0.02 &0.08 &0.63 &0.63 &0.62 &0.62 &0.66 &0.07 \\
&Total		&2.91	&6.05	&9.94	&5.19	&5.64	&28.68	&\textbf{1.09}	&1.56	&14.14	&14.14	&14.14	&14.13	&14.17	&\textbf{1.3}\Bstrut\\
     \hline
\multicolumn{2}{r}{Accuracy} &99.3	&99.1	&99.1	&99.1	&99.0	&99.1	&98.6	&99.3	&99.1	&99.1	&99.1	&99.0	&99.1	&98.6 \Tstrut\Bstrut\\     
\hline    
    \end{tabular}}
\label{t:Compare1}
\end{table*}

The state $s_t$ can be expressed by Equation~\ref{e:st}. Here $s_t$ is the set containing all the quantization depth $Q$, the pruning remaining amount $P$, and the reward $r$ from step $t-\tau$ to step $t$. It also contains $t$, the index of current step. We want the state of the environment to well reflect the history of the optimization process. Hence, the state contains the values of $Q$ and $P$ in previous $\tau$ steps. To guarantee that the state set has the same dimension at any optimization step, we have $Q_{t-\tau}=Q_{0}$ and $P_{t-\tau}=P_{0}$ if $t$ is less than $\tau$.
\begin{equation}
r_t= (\alpha_t/\alpha_{t-1})^\lambda\cdot \beta_{t-1}/\beta_t
\label{e:rt}
\end{equation}

The reward $r_t$ can be expressed by Equation~\ref{e:rt}. Here, $\alpha_t$ and $\alpha_{t-1}$ are the accuracy at current step $t$ and previous step $t-1$, respectively. $\beta_t$ and $\beta_{t-1}$ are the energy consumption at step $t$ and step $t-1$, respectively. In this paper, we target on the energy consumption and the accuracy only. Since the area overhead is highly correlated with energy consumption, EDC could also improve the area efficiency of edge devices effectively. Intuitively, decreasing the quantization depth and the pruning remaining amount would reduce the energy consumption and at the same time decrease the accuracy. The reinforcement learning algorithms can automatically find the trade-off point between the accuracy $\alpha$ and the energy consumption $\beta$. Experiments results show that the multiplication of energy and accuracy as the reward is better that the summation of these two metrics. We use a third parameter $\lambda$ to show the importance of accuracy over the energy consumption. It is normally greater than $1$, and is fixed during the optimization. We test different values of $\lambda$ in experiments, and find that $\lambda=3$ is an optimal value.

\section{Experiments}

\textbf{Algorithm setup:} we use a state-of-the-art reinforcement learning algorithms SAC (soft actor-critic)~\cite{haarnoja2018soft} to train our optimization model. Compared with classical large-space problems, the searching space in our problem is not large, and SAC can approach the optimal solutions very quickly (less than ONE day on ImageNet using a single graphic card Titan Xp). We test EDCompress on the ImageNet, CIFAR-10 and MNIST datasets using three different neural networks: VGG-16~\cite{simonyan2014very}, MobileNet~\cite{howard2017mobilenets} and LeNet-5~\cite{lecun1998gradient}. These three models are different in characteristics. VGG is a complex deep neural network. MobileNet is designed for computation efficiency. LeNet-5 is a simple neural network with only two neural layers. We study four dataflow types, which are the most commonly used dataflow types. In each episode, we start from a well-train model. When the last episode ends, we restore the weights from a saved checkpoint, and reset the quantization depth/pruning remaining amount in each layer.

\noindent\textbf{Hardware setup:} we implement popular dataflows $X:Y$, $F_X:F_Y$, $X:F_X$ and $C_I:C_O$ on the Xilinx Virtex UltraScale FPGA, and obtain the energy consumption and area overhead from the Xilinx XPE toolkit~\cite{xilinx2}, which can be reported in a few seconds. In the logic part, the multipliers and adders are implemented on LUTs (lookup tables). An $M\times N$ multiplier requires $M/2 \times (N+1)$ LUTs~\cite{walters2016array}. In our experiment, parameters in the feature map are quantized by $10$ bits, while the weights are quantized by $q$ bits ($q$ ranging from 0 to 8). Hence, we need $5q$ LUTs for a single $10 \times (q+1)$ multiplier. In the memory part, the on-chip memory is implemented on RAM (Random-Access Memory) modules. During inference, to save the memory space, the input feature map is not kept after the computation of each layer. Hence, the size of the memory modules must support the weights in all layers plus the maximum feature map in the model.

\begin{figure*}[!t]
  \centering
  \includegraphics[width=1.0\textwidth]{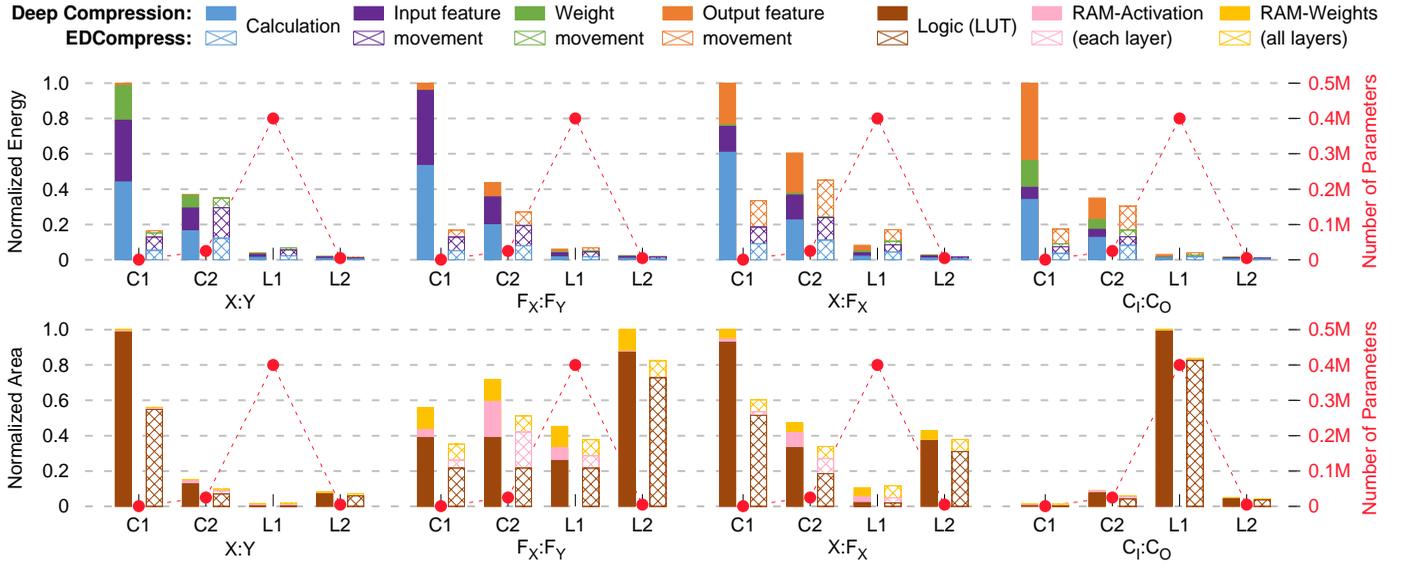}\\
  \caption{Layerwise comparison of energy consumption and area overhead between EDCompress and Deep Compression on LeNet-5. The color bar denotes the breakdown of energy and area, and the red polyline denotes the number of parameters in each layer (right-hand y-axis)}
  \label{f:Compare}
\end{figure*}

\begin{figure*}[!t]
  \centering
  \includegraphics[width=1.0\textwidth]{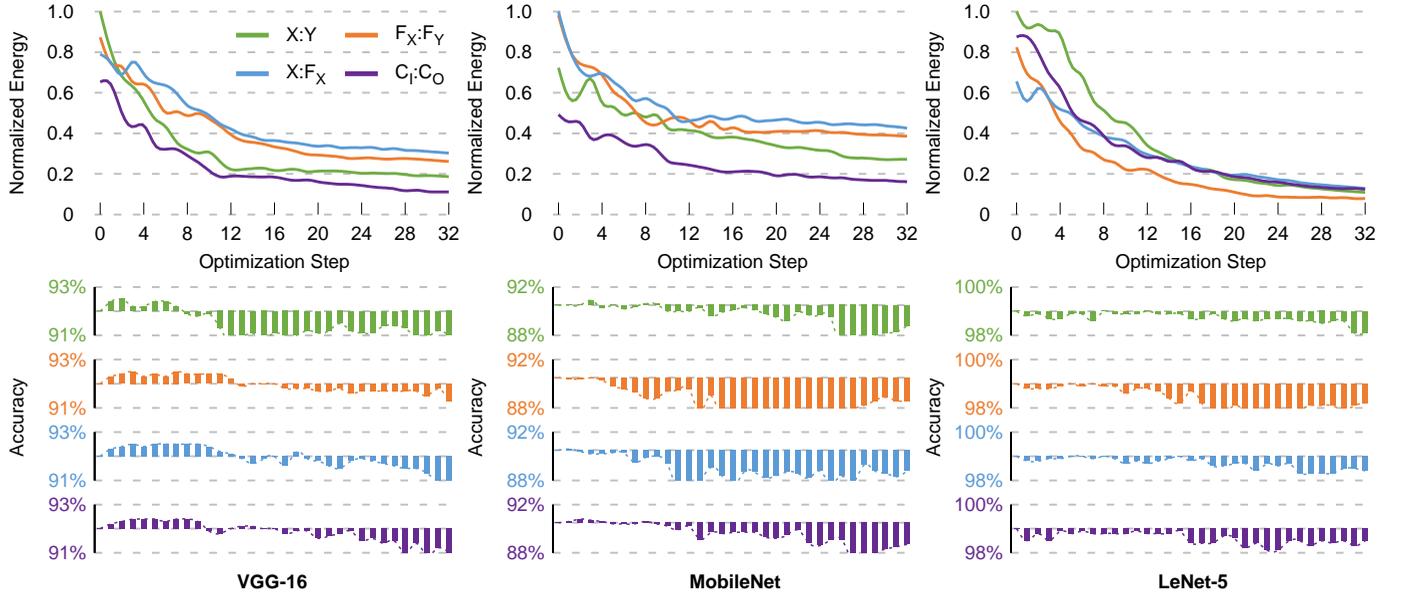}\\
  \caption{Optimization process of EDCompress on CIFAR-10 (VGG-16/MobileNet) and MNIST (LeNet-5). In each episode, we run thirty-two steps. The curves show the energy consumption of four dataflows, and the bars show the accuracy of the model}
  \label{f:Time}
\end{figure*}

\subsection{Comparison with the State-of-the-Art}

EDCompress is effective on all kinds of datasets. Table~\ref{t:Compare21}, Table~\ref{t:Compare22} and Table~\ref{t:Compare1} compare EDCompress with the state-of-the-art work on the ImageNet, CIFAR-10 and MNIST datasets. Compared with HAQ on ImageNet, our EDCompress test on four dataflow types and could achieve averaged 3.8X, and 3.9X improvements on energy and area efficiency with similar accuracy.
In this paper, we focused on small-size datasets because we are targeting on edge devices running lite applications. It shows that among the four dataflows, EDCompress could more effectively reduce the energy consumption and area overhead, with negligible loss of accuracy. Compared with the state-of-the-art work, EDCompress shows 9X improvement on energy efficiency and 8X improvement on area efficiency in LeNet-5, in average of the four dataflow types. It also shows 11X/6X improvement on energy/area efficiency in VGG-16. If we optimize the model by EDCompress, the dataflow $F_X:F_Y$ is the most appropriate choice for LeNet-5 in terms of energy consumption and area overhead, and the dataflow $X:Y$ is the most appropriate one for VGG-16.

Comparisons also indicate that instead of compressing the model size, EDCompress is more efficient in the reduction of energy consumption and area overhead. For example, in Figure~\ref{f:Compare}, we compare the energy and area between EDCompress and Deep Compression (DC)~\cite{han2015deep}, layer by layer. From the figure, EDCompress shows 2.4X higher energy efficiency and 1.4X higher area efficiency than DC. We can see that in the third layer, DC shows better performance than EDCompress on energy consumption because this layer contains 93\% of the total parameters. However, this layer does not contribute to most of the energy consumption. In fact, compressing the first layer would be more helpful on the energy reduction, although it only contains 0.1\% of the parameters. Figure~\ref{f:Compare} and Table~\ref{t:Compare1} show that EDCompress can reduce much more energy consumption and area overhead in the first layer, compared with previous work. Another example is the dataflow $C_I:C_O$, whose third layer contributes to most of the area overhead. From the figure, we can see that EDCompress shows higher area efficiency than DC in the third layer. This observation further prove that EDCompress is more efficient in the reduction of hardware resources.

\begin{figure*}[!t]
  \centering
  \includegraphics[width=1.0\textwidth]{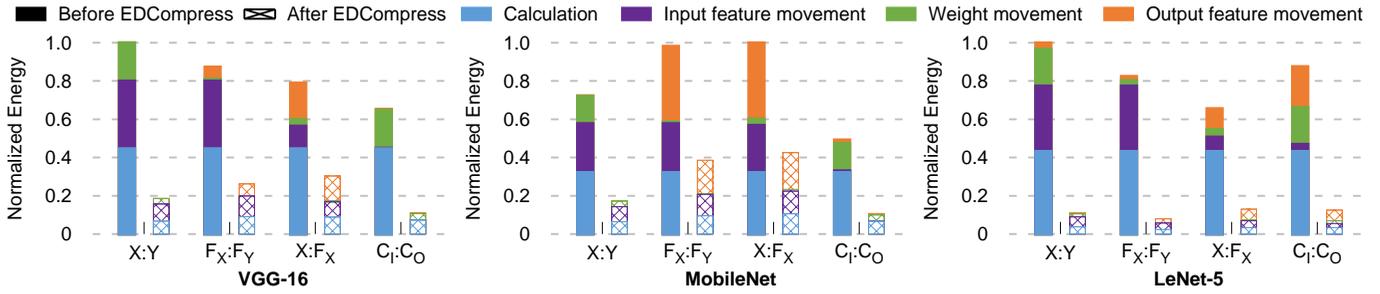}\\
  \caption{Energy consumption breakdown before and after the optimization of EDCompress. The solid bar and patterned bar represent results before and after the EDCompress, respectively}
  \label{f:EnergyBreak}
\end{figure*}

\begin{figure*}[!t]
  \centering
  \includegraphics[width=1.0\textwidth]{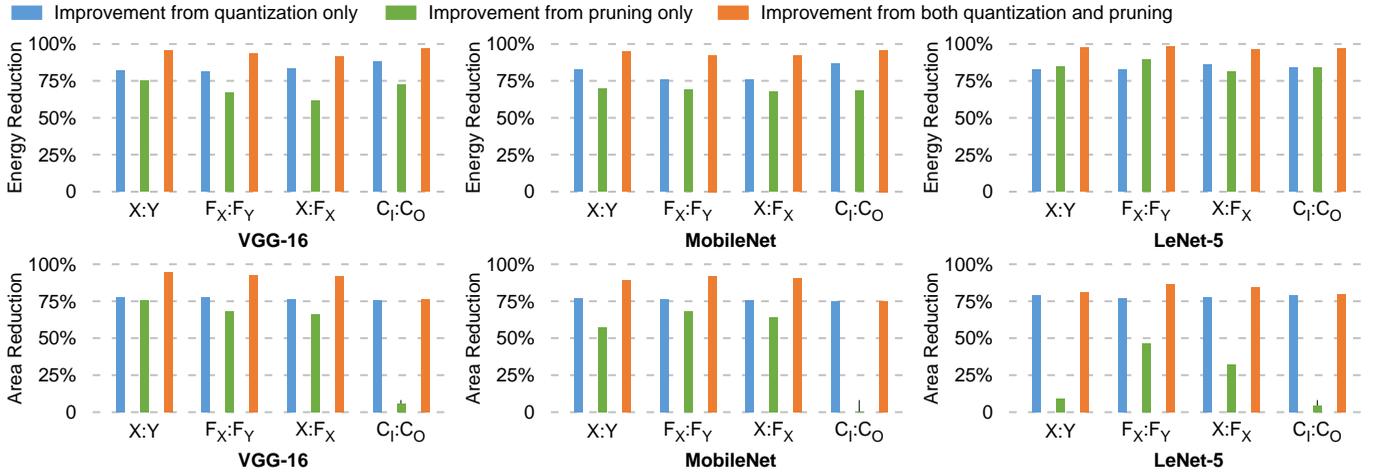}\\
  \caption{The performance of EDCompress by applying quantization technique only, pruning technique only, and both quantization/pruning techniques}
  \label{f:contribute}
\end{figure*}

\subsection{Insights on Dataflow}

Quantization and pruning have different effects on different dataflow designs. Figure~\ref{f:Time} shows the optimization process of the hardware accelerators using three neural networks in terms of energy consumption and accuracy. We start the optimization from a model with activations quantized in 10-bit and weights quantized in 8-bit. From the figure, we can see that the reinforcement learning algorithm could effectively reduce the energy consumption, with negligible loss of accuracy. Figure~\ref{f:EnergyBreak} shows the energy consumption breakdown of each dataflow before EDCompress (model using 10-bit activations and 8-bit weights) and after EDCompress. If we compare the optimized result from EDCompress with the original model, the energy efficiency in VGG-16, MobileNet, LeNet-5 networks can be improved by 20X, 17X, 37X, respective. More specifically, around 55\% energy consumption is saved from processing elements and the rest 45\% are saved from data movement.

The results also indicate that optimization could change our choice on dataflow types. Those dataflows that do not show good energy efficiency before the optimization may show very high energy efficiency after the optimization. Take the VGG-16 for example, before the optimization, the dataflow $X:Y$ consumes the most energy among the four dataflows. However, after the optimization, $X:Y$ consumes the second lowest energy consumption. This is because the energy consumption of hardware accelerators includes the energy of MAC operations on processing elements, and the energy on data movement. As we can see from Figure~\ref{f:EnergyBreak}, given the fixed pruning remaining amount and quantization depth, the energy consumed on processing elements are almost the same. The efficient way to save the energy is to spent less energy on data movement. Due to the optimization, the energy consumed on data movement decreases because the amount of delivered data is reduced. In this process, different dataflow designs have different amount of reduction on the delivered data. $X:Y$, in this case, is more efficient in data movement reduction, and therefore we can save more energy consumption on this dataflow than other dataflow types.

\subsection{Insights on Quantization/Pruning}

The effectiveness of quantization and pruning techniques on the reduction of energy consumption and area overhead is highly related to the dataflow type. Figure~\ref{f:contribute} shows their individual contributions. From the figure, we can see that in most cases, both quantization and pruning can effectively reduce the energy consumption and area overhead. More specifically, if we apply quantization technique only, EDCompress can achieve 5.6X improvement on energy efficiency and 4.3X improvement on area efficiency. If we apply pruning techniques only, EDCompress can achieve 3.8X/1.7X improvements on energy/area efficiency.

We have two observations in Figure~\ref{f:contribute}. First, pruning shows very little improvement on area overhead of the $C_I:C_O$ dataflow design. Second, the small-scale model LeNet-5 prefers quantization over pruning. This is because in these cases, the accelerator demands more area on the processing elements than the memory modules. Pruning can effectively reduce the area of memory modules because of the reduction of model size. However, it is not good at decreasing the area of processing elements. Quantization, on the other hand, could reduce the area of both processing elements and memory modules effectively. Hence, the quantization technique would be more useful in these cases.

\section{Conclusions}

We propose EDCompress, an energy-aware model compression method for dataflows. To the best of our knowledge, this is the first paper studying model compression problem with the knowledge of the dataflow design in accelerators. Considering the very nature of model compression procedures, we recast the optimization to a multi-step problem, and solve it by the reinforcement learning algorithm. Experiments show that EDCompress could improve 20X, 17X, 37X energy efficiency in VGG-16, MobileNet, LeNet-5 networks, respectively, with negligible loss of accuracy. EDCompress could also find the optimal dataflow type for specific neural networks, which can guide the deployment of CNN on hardware systems. However, deciding which dataflow type to use in the edge device depends on many other constraints, such as the expected computation speed, the thermal design power, the fabrication budget, etc. Therefore, we leave the final decision to hardware developers. 

\bibliography{main}
\bibliographystyle{IEEEtran}

\end{document}